# Atypical Facial Landmark Localisation with Stacked Hourglass Networks: A Study on 3D Facial Modelling for Medical Diagnosis

**Gary Storey[1], Ahmed Bouridane[2], Richard Jiang*[3] and Chang-tsun Li[4]**

**Abstract –** While facial biometrics has been widely used for identification purpose, it has recently been researched as medical biometrics for a range of diseases. In this chapter, we investigate the facial landmark detection for atypical 3D facial modelling in facial palsy cases, while potentially such modelling can assist the medical diagnosis using atypical facial features. In our work, a study of landmarks localisation methods such as stacked hourglass networks is conducted and evaluated to ascertain their accuracy when presented with unseen atypical faces. The evaluation highlights that the state-of-the-art stacked hourglass architecture outperforms other traditional methods.

## 1. Introduction

The task of landmark localisation is well established within the domain of computer vision and widely applied within a variety of biometric systems. Biometric systems for person identification commonly apply facial [13, 31-36], ear [28] and hand [26] landmark localisation, where Fig.1 shows example of these landmark localisation variations. The landmark localisation task can be described as predicting *n* fiducial landmarks when given a target image, the human face is one common target for landmark localisation

[1] Computer and Information Sciences, Northumbria University, Newcastle upon Tyne, NE1 8ST, UK, e-mail : gary.storey@northumbria.ac.uk

[2] Computer and Information Sciences, Northumbria University, Newcastle upon Tyne, NE1 8ST, UK, e-mail: ahmed.bouridane@northumbria.ac.uk

[3] Computing & Communication, Lancaster University, Lancaster, LA1 4WY, UK, e-mail : r.jiang2@lancaster.ac.uk

[4] School of Info Technology, Deakin University, Waurn Ponds VIC 3216, Australia, e-mail : changtsun.li@deakin.edu.au

where semantically meaningful facial landmarks such as the eyes, nose, mouth and jaw line are predicted. The purpose of the landmark localisation task within biometric system pipeline is to aid the feature extraction process from which identification can be predicted. Generally there are two types of features extracted these being geometry-based and texture features, geometry-based features use the landmarks locations directly as features for example ratio distances between these landmarks [13]. Texture features instead use the predicted landmarks as local guides for feature extraction from specific facial locations. It is key that the landmark localisation performed is accurate in order to reduce poor feature extraction and therefore potential system errors.

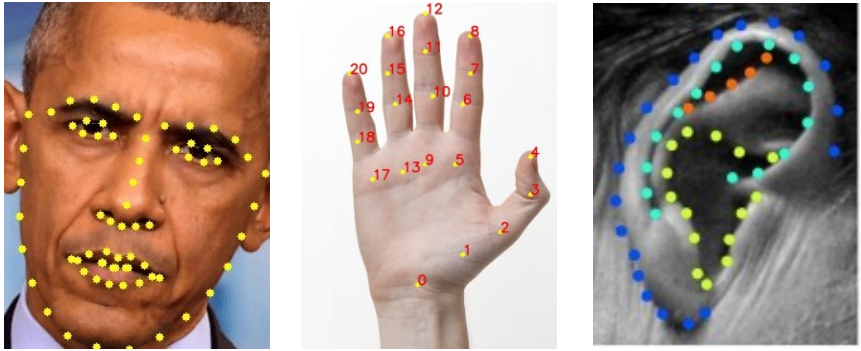

**Figure 1.** Landmark Localisation application examples: (Left) - Face, (Centre) - Palm, (Right) - Ear.

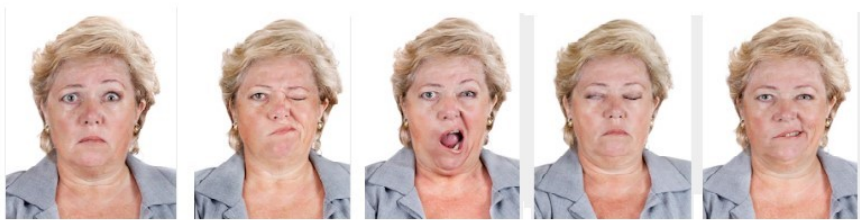

**Figure 2.** Asymmetrical face examples.

The main focus of this chapter is facial landmark localisation which has a long history of research and is also referred to as face alignment. Research to date can be generally divided into three categories.

Holistic based approaches such as Active Appearance Models (AAMs) [7, 11] solve the face alignment problem by jointly modelling appearance and shape. Local expert based methods such as Constrained Local Models (CLMs) [9] learn a set of local experts detectors or regressors [24, 19] and apply shape models to constrain these. The most recent advancements which have attained state-of-the-art results apply CNN based architectures with probabilistic landmark locations in the form of heat maps [3]. While these advancements have increased both the accuracy and reduced the computational time of the landmark localisation process challenges still exist. One specific challenge is that of asymmetrical faces [21], while a majority of the population have typical face structures with small degrees of asymmetry, as shown in Fig.2, there exists a section who for a variety of reasons including illness and injury display atypical facial structure, including those with a large degree of asymmetry. To enable biometric systems that are universally accessible and do not discriminate against those with atypical face structures due to poor feature extraction, it important to ascertain the accuracy of landmark localisation methods on this type of facial structure, especially as the public training sets do not contain specific samples of this demographic.

In this chapter a study is presented, which evaluates the accuracy of a number of landmark localisation methods, namely on with two data sets containing atypical faces. A specific focus on the state-of-the-art stacked hourglass architecture is also documented. The remaining sections of this chapter are structured as follows, firstly a brief history of landmark localisation methods is presented in section 2. Section 3 provides a detailed overview of the stacked hourglass architecture in general and the Face Alignment Network (FAN) method [2] applied specifically for facial landmark localisation. The evaluation is presented in section 4, which highlights the accuracy of each method against the data sets. Finally section 5 provides a conclusion to this chapter and explores future areas of research.

## 2. Landmark Localisation History

In this section a brief description of historically important landmark localisation methods is presented. The first subsection details non-deep learning based methods which up until recent years were considered state-of-the-art, while the second subsection concentrates on the deep learning based methods from recent literature.

## 2.1 Traditional Methods

Within the traditional methods the Active Shape Model (ASM) developed by [8] provided one of the first great breakthrough methods which could be applied to landmark localisation, they followed up this work an alternative method namely AAMs [7]. Both methods while not specifically designed for face landmark localisation leverage the idea of defining statistically developed deform-able models. There are similarities and distinct differences between the methods, while both use a statically generated model consisting of both texture and shape components learnt from a training data set, the texture component and how it is applied in the landmark fitting process are distinct to each method. The shape model is composed through the alignment of the training images by using a variation of the Procustes method which scales, rotates and translates the training shapes so that they are aligned as closely as possible. Principle Component Analysis (PCA) is then carried out on the training images reducing the dimensions of the features while retaining the variance in the shape data. A mean shape is also generated which is often used as a starting point for fitting to new images. The ASM is considered to be in the CLM group of methods, these model types use the texture model as local experts in which they are trained on texture information taken from a small area around each landmark. The local expert ASM uses a small set of gray-scale pixel values perpendicular to each landmark while other CLM techniques use a block of pixels around the landmarks or other feature descriptors such as SIFT [9, 16]. The fitting of the model is carried out via the optimisation of an objective function using the prior shape and the sum of the local experts to guide the alignment process. AAM differs from the CLM group of methods by using a texture model of the entire face rather than regions. To create this all face textures from the training images are warped to a mean-shape, transformed to grey scale and normalised to reduce global lighting effects. PCA is then applied to create the texture features. Alignment on an unseen image is carried out by minimising the difference between the textures of the model and the unseen image [7].

Further advancements in accurate and computationally efficient landmark localisation arrived with the application of regression based fitting methods rather than sliding-windows based approaches. Regressors also provide detailed information regarding the local texture prediction criteria when compared with the classifier approach which is a binary prediction of match or not. [24] proposed a method named Boosted Regression coupled with Markov Networks, in which they apply Support Vector Regression and local appearance based features to predict 22 initial facial landmarks in an iterate manner, Markov Networks are then used to sample new facial locations to apply the regressor to in the next iteration. Cascaded regression was then applied by [10, 29] in which a cascade of weak regressors is applied to reduce the alignment error progressively while providing computationally efficient regression methods. Different feature types have been applied these for example [6] have recently produced a face alignment method based up a multilevel regression using fern and boosting. This has been subsequently built upon in [5] where a regression based technique named Robust Cascaded Posed Regression, which can also differentiate between landmarks that are visible and non-visible (occlusion) and estimate those facial landmarks that may be covered by another object such as hair or a hand proposed. [19] have also applied a regression technique with local binary features and random forests to produce a technique that is both accurate and computationally inexpensive meaning that the algorithm can perform at 3000fps for a desktop PC and up to 300fps for mobile devices.

The previous methods predicted facial landmarks on faces in limited poses at most between ±60 degrees, both the Tree Shape Model (TSM) [18] and PIFA [12] are notable methods which could handle a greater range of face pose. The TSM [18] was unique amongst landmark localisation methods in that it did not use a regression or iterative methods for determining landmarks positions, instead this used the HOG parts to determine location based upon appearance and the configuration of all parts was scored to determine the best fit for a face. The final X and Y coordinates of the predicted landmarks are derived from the centre of a bounding box for that specific parts detection. [12] proposed PIFA as a significant improvement in dealing with all face poses and determining the visibility of a landmark across poses for up to 21 facial landmarks. This method extended 2D cascaded landmark localisation through the training of two regressors at each layer of the cascade. The first regressor predicts the update for the camera projection matrix which map to the pose angle of the face.

The second is responsible for updating the 3D shape parameter which determines 3D landmarks positions. Using 3D surface normal's, visibility estimates are made based upon a z coordinate, finally the 3D landmarks are then projected to the 2D plane.

## 2.2 Deep Learning Methods

The initial deep learning Convolutional Neural Network (CNN) based landmark localisation methods while displaying high accuracy were limited to a very small set of sparse landmarks when compared with previous traditional methods. A Deep Convolutional Network Cascade was proposed in [23], this consisted of a 3 stage process for landmark localisation refinement, at each level of the cascade multiple CNNs were applied to predict the locations for individual and subsets of the landmarks. This method only considered 5 landmarks and the capability to expand this to further landmarks is computationally expensive due to the nature of using individual CNNs to predict each landmark. [27] applied multi-task learning to enhancement in which they trained a single CNN with not only facial landmark locations but also gender, smile, glasses and pose information. Linear and logistic regression were used to predict the values for each task from shared CNN features. When directly compared with the Deep Convolutional Network Cascade [23] they showed increased landmark accuracy with a significant computational advantage of using a single CNN. A Backbone-Branches Architecture was applied in [15] which outperformed the previous methods in terms of both accuracy and speed for 5 facial landmarks. This model consisted of a multiple CNNs, a main backbone network which generates low-resolution response maps that identify approximate landmark locations, then branch networks produce fine response maps over local regions for more accurate landmark localisation.

The next generation of deep learning methods expanded on these initial methods increasing the number of landmarks detected to the commonly used 68. HyperFace applies a multi-task approach which also considered face detection. The idea of the multi-task approach is that inter-related tasks can strengthen feature learning and remove over-fitting to a single objective. HyperFace used a single CNN originally AlexNet, but modified this by taking features from layers 1, 3 and 5, concatenating these into a single feature set, then passing these through a further convolutional layer prior to the fully connected layers for each task. At the same time the fully-convolutional network (FCN) [15] emerged as a technique, in which rather than applying regression methods to predict landmarks coordinates, they are based upon response maps with spatial equivalence to the raw images input. Convolutional and de-convolutional networks are used to generate a response map for each facial landmark, further localisation refinement applying regression was then used in [14, 25, 4]. The stacked hourglass model proposed in [17] for human pose estimation which applied repeated bottom-up then top-down processing with intermediate supervision has been applied to the landmark localisation in a method called the FAN [3] this has shown state-of-the-art performance on a number of evaluation data sets. Further more this method expanded the capability of detection from 2D to 3D landmarks through the addition of a depth predictions CNN which takes a set of predicted 2D landmarks and generates the depth. At the time of publication the FAN method outperformed previous methods for accurate landmark localisation.

## 3. Stacked Hourglass Architecture

In this section a detailed overview of the stacked hourglass architecture is given [17]. This architecture has proven to be extremely accurate for landmarks localisation tasks in both human pose detection where landmarks include the head, knee, foot and hand, and also for facial landmark localisation [17, 2]. The capability to and potential to generalise well to other types of landmark localisation.

### 3.1 Hourglass Design

The importance of capturing information at every scale across an image was the primary motivation for [17] design of the hourglass network. Originally designed for the task of human pose estimation where the key components of the human body such as head, hands and elbow are best identified at different scales. The design of the hourglass provides the capability to capture these features across different scales and bring these together in the output as pixel-wise predictions. The name hourglass is taken from the appearance of the networks down sampling and up sampling layers which are shown in Fig.3. Given an input image to the hourglass, the network initially consists of down sampling convolutional and max pooling layers which are used to predict features down to a very low resolution. During this down sampling of the input the network branches off prior to each max pooling step and further convolutions are applied on the pre-pooled branches, this is then fed back into the network during up sampling. The purpose of network branching is to capture intermediate features across scales, without the application of these branches rather than learn features at each scale the network would behave in a manner previously shown in Fig.2 where initial layer learn general features and deeper layers learn more task specific information. Following the lowest level of convolution the network then begins to up sample back to the original image resolution through the application of nearest neighbour up sampling and element wise addition of the previously branched features. Each of the cuboids in Fig.3 is a residual module also known as bottleneck blocks as shown in Fig.4. These blocks are the same as those used within the ResNet architectures.

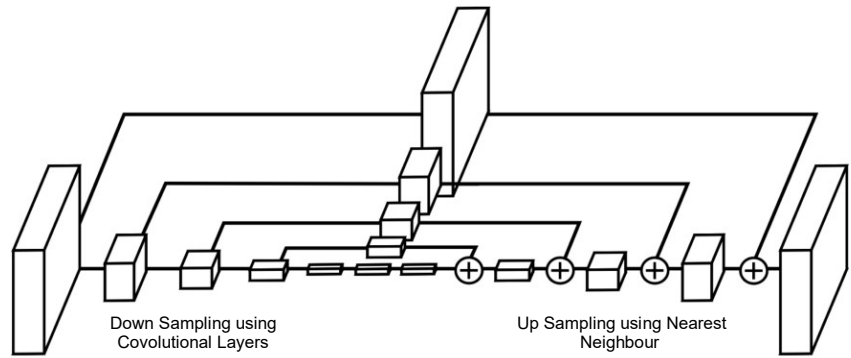

**Figure 3.** Hourglass Design.

## 3.2 Stacked Hourglass with Intermediate Supervision

The final architecture proposed by [17] took the hourglass design and stacked $n$ hourglasses in an end-to-end fashion, where in the best performing configuration for human pose estimation was $n = 8$. Each of these hourglass's is independent in terms of the weight parameters. The purpose of this stacked approach it to provide a mechanism in which the predictions derived from a single hourglass can be evaluated at multiple stages within the total network. A key technique in the use of this stacked design is that of intermediate supervision, in which at the end of each individual hourglass a heat-map output is generated to which a Mean Square Error (MSE) loss function can be applied. This process is similar to the iterative processes found in other landmark localisation methods, where each hourglass further refines the features and therefor the predictions as they move through the network. Following the intermediate supervision the heat-map, intermediate features from the hourglass and also the feature from the previous hourglass are added. To do this a $1 \times 1$ convolutional layer is applied to remap the heat-map back into feature space.

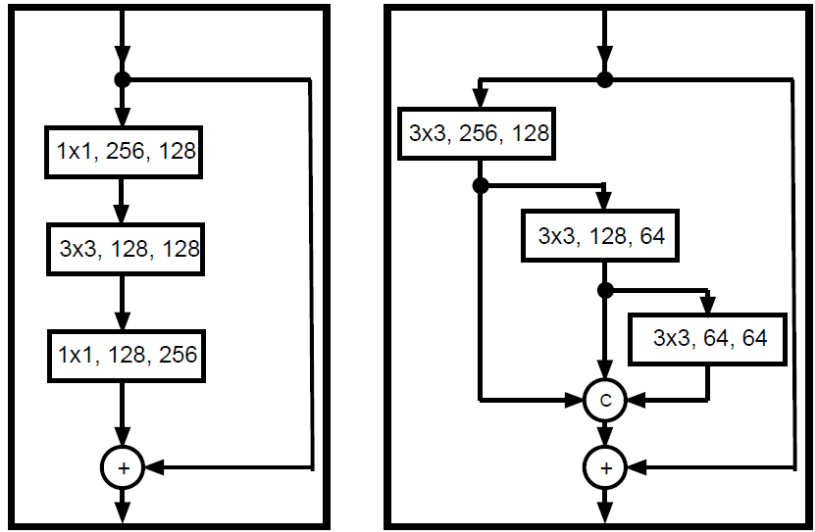

**Figure 4.** Block Design: (Left) The basic bottleneck block. (Right) The hierarchical, parallel and multi-scale block of FAN.

### 3.3 Facial Alignment Network

The FAN takes the stacked-hourglass design and trains this for the task of facial landmark localisation. Landmark localisation has similar challenges to that of human pose estimation, where the face landmarks are represented at different local scales within the context of the global context human face. Architectural changes are made to the network design where FAN reduces the total number of stacked Hourglass's from 8 to 4. Also the structure of the convolutional blocks are changed from bottle necks to a hierarchical, parallel and multi-scale block, which performs three levels of parallel convolution alongside batch normalisation before outputting the concatenated feature map (Fig.4). It was shown in [2] that when total parameter number is equal this block type outperforms the bottleneck design. The parameters of the $1 \times 1$ convolutional layers are changed to output heat-maps of dimension $H \times W \times m$, where $H$ and $W$ are the height and width of the input volume and $m$ is the total number of facial landmarks predicted where $m = 68$.

Training of the FAN was completed using a synthetically expanded version of the 300-W [20] named the 300-W-LP [30], while the original 300-W was also used to fine-tune the network. Data augmentation was applied during training, this employed random flipping, rotation, colour jittering, scale noise and random occlusion. The training applied a learning rate of $10^{-4}$ with a mini-batch size of 10. At 15 epoch intervals the learning rate was reduced to $10^{-5}$ then again to $10^{-6}$. A total of 40 epochs were used to fully train the network. The MSE loss function is used to train the network:

$$MSE = \frac{1}{n}\sum(Y_i - \hat{Y}_i)^2 \quad (1)$$

where $Y_i$ is predicted heat-map for the $i^{th}$ landmark and $\hat{Y}_i$ is a ground truth heat-map consisting of a 2D Gaussian centred on the landmark location of the $i^{th}$ landmark.

### 3.4 Depth Network for 3D landmarks

A further extension to the FAN method is the capability to extend the 2D facial landmarks to 3D, this is achieved through the application of a second network. This second network takes as the input the predicted heat maps from the original 2D landmark localisation and the face image. The heatmaps guide the networks focus on areas of the image at which depth should be predicted from. This network is not hourglass based but instead a adapted ResNet-152, where the input takes $3 + N$ where 3 is the RGB channels of the image and $N$ is the heatmap data where $N = 68$. The output of the network is $N \times 1$. Training applied 50 epochs using similar data augmentation as the 2D model training, with a learning rate of $10^{-3}$ and an L2 loss function.

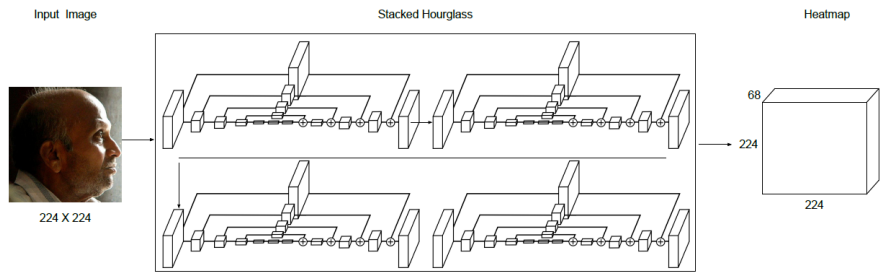

**Figure 5.** Facial Alignment Network Architecture Overview.

## 4. Evalaution

Within this section an evaluation on landmark localisation. The evaluation was conducted using PyTorch 0.4 on Windows 10 with a Nvidia GTX 1080 GPU.

A key foundation for many end-to-end automated diagnostic pipelines is the requirement to have precise facial landmark localisation. It is common practice to use these detected facial landmarks directly as geometric features or as indicators of areas of interest from which feature extraction can occur. In previous research [22] it has been highlighted that a number of methods that have gained state-of-the-art accuracy on symmetrical faces do not display the same level of accuracy when the face displays asymmetry, like those diagnosed with facial palsy.

In this study we expand the previous research to include a larger sample size, while also investigating the impact new deep learning methods have in comparison with the previous landmark localisation methods. The methods evaluated in order of publication are the Tree

Shape Model (TSM) [18], the DRMF [1] and the deep learning based Face Alignment Network (FAN) [3].

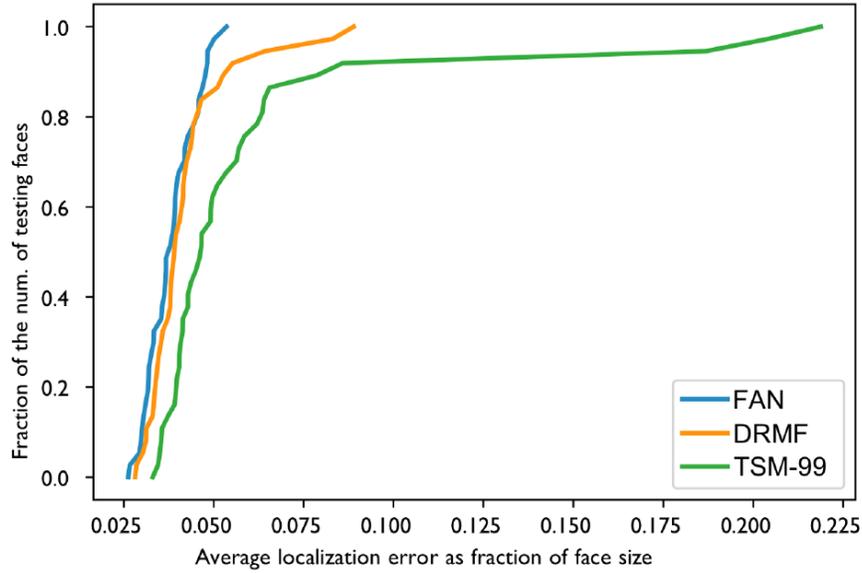

**Figure 6.** Cumulative localisation error distribution from Facial Palsy test set A.

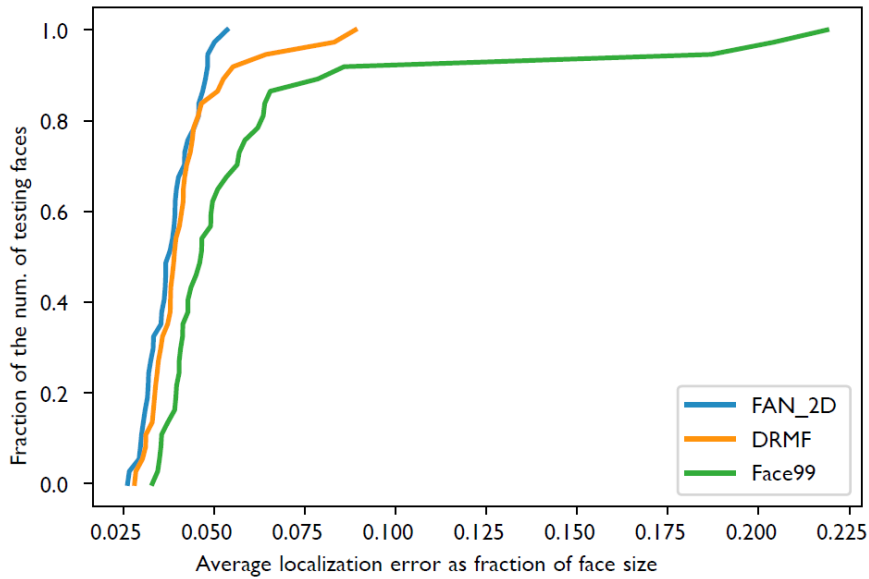

**Figure 7.** Cumulative localisation error distribution from Facial Palsy test set B.

The evaluation of landmark localisation accuracy uses two separate data sets both containing images of individuals with varying grades of facial palsy. Data set A consists of 47 facial images which have 12 ground truth landmarks. Data set B consists of a further 40 images which are annotated with 18 ground truth landmarks per image. Normalised Mean Error (NME) using face size normalisation as described in [3] is used as the evaluation metric. Different methods of landmark localisation have variance in both the number and specific locations of the landmarks predicted, a subset of facial landmarks are used which are common across all methods which allows for a comparative analysis.

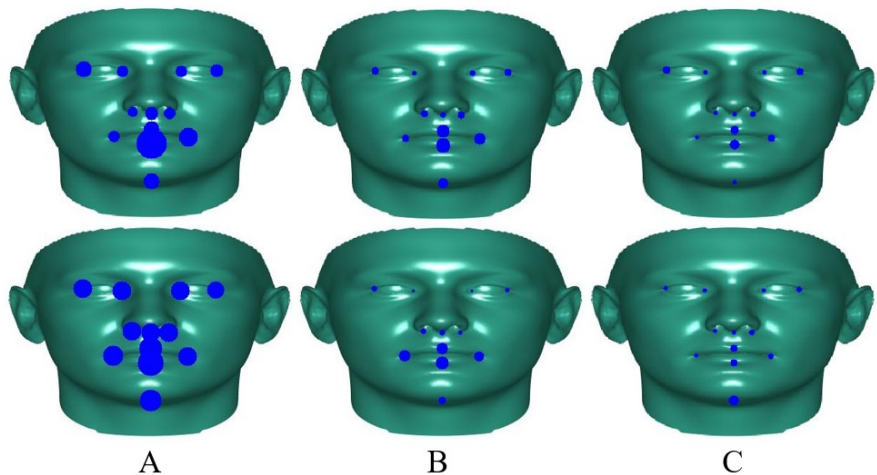

         A                    B                    C

**Figure 8.** Normalised Mean Error Per Landmark: (Top) - Facial Palsy Test Set A (Bottom) – Facial Palsy Test Set B, (A) - TSM 99 Part Shared, (B) - DRMF, (C) – FAN.

The cumulative localisation NME error for data sets A and B are shown in Fig.6 and Fig.7 respectively. The results show that the deep learning based FAN method displays a consistently higher level of accuracy across both datasets. DRMF performs accurate landmark prediction for certain test samples but specifically in test set B where there is high degree of facial asymmetry there is a percentage of the sample for which the error increases by a substantial amount. Finally TSM performs poorly in general comparatively and this error grows substantially as the level of facial asymmetry increases. Analysing the prediction NME error for a specific selection of landmarks as shown in Fig.8, the results show that while FAN and DRMF have similar level of accuracy for the eye and nose landmarks, the mouth which has the largest range of asymmetrical deformation is where the deep learning based FAN excels. Fig.9 provides a visual example of the landmark localisation output, this highlights the capability of the deep learning FAN method to provide a high level of accuracy when fitting landmarks to the face and specifically the mouth region when compared with previous techniques.

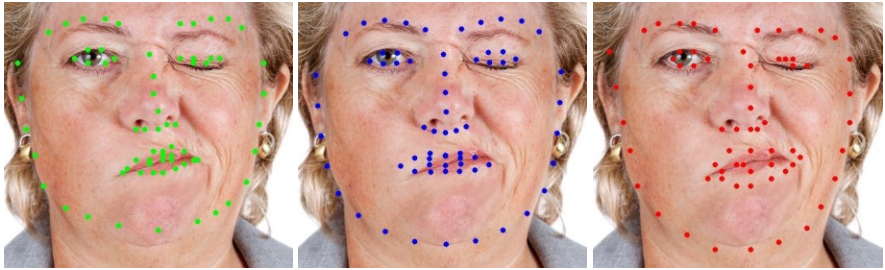

**Figure 9.** Landmark Localisation fitting example for each evaluated method. (Left) - FAN, (Centre) DRMF, (Right) - TSM.

## 5 Conclusion

The focus of this chapter was to study how accurately current landmark localisation methods predict landmarks on atypical faces. It was found that of the methods evaluated only the state-of-the-art FAN method could accurately predict facial landmarks, especially on the difficult mouth landmarks which show a higher degree of atypical appearance. The stacked hourglass architecture and it's derivative the FAN, prove to be a high performing method for landmark localisation, which has the potential to be applied to other landmark localisation tasks such as the ear and hand.

Atypical Facial Landmark Localisation: A study                                     15[3] Adrian Bulat and Georgios Tzimiropoulos. How far are we from solving the 2D & 3D Face Alignment problem? (and a dataset of 230,000 3D facial landmarks). *2017 IEEE International Conference on Computer Vision (ICCV)*, pages 1021–1030, 10 2017.

[4] Adrian Bulat and Yorgos Tzimiropoulos. Convolutional aggregation of local evidence for large pose face alignment. In *Procedings of the British Machine Vision Conference 2016*, pages 1–86. British Machine Vision Association, 2016.

[5] Xavier P. Burgos-Artizzu, Pietro Perona, and Piotr Dollar. Robust Face Landmark Estimation under Occlusion. In *2013 IEEE International Conference on Computer Vision*, pages 1513–1520. IEEE, 12 2013.

[6] Xudong Cao, Yichen Wei, Fang Wen, and Jian Sun. Face alignment by explicit shape regression. *International Journal of Computer Vision*, 107(2):177–190, 2014.

[7] T.F. Cootes, G.J. Edwards, and C.J. Taylor. Active appearance models. *IEEE Transactions on Pattern Analysis and Machine Intelligence*, 23(6):681–685, 2001.

[8] Tim Cootes, Er Baldock, and J Graham. An introduction to active shape models. *Image Processing and Analysis*, pages 223–248, 2000.

[9] David Cristinacce and Tim Cootes. Automatic feature localisation with constrained local models. *Pattern Recognition*, 41(10):3054–3067, 2008.

[10] Piotr Dollár, Peter Welinder, and Pietro Perona. Cascaded pose regression. In *Proceedings of the IEEE Computer Society Conference on Computer Vision and Pattern Recognition*, pages 1078–1085, 2010.

[11] Ralph Gross, Iain Matthews, and Simon Baker. Active appearance models with occlusion. *Image and Vision Computing*, 24(6):593–604, 2006.

[12] Amin Jourabloo and Xiaoming Liu. Pose-Invariant Face Alignment via CNN-Based Dense 3D Model Fitting. *International Journal of Computer Vision*, 124(2), 2017.

[13] Aniwat Juhong and C. Pintavirooj. Face recognition based on facial landmark detection. In *2017 10th Biomedical Engineering International Conference (BMEiCON)*, pages 1–4. IEEE, 8 2017.

[14] Hanjiang Lai, Shengtao Xiao, Yan Pan, Zhen Cui, Jiashi Feng, Chunyan Xu, Jian Yin, and Shuicheng Yan. Deep Recurrent Regression for Facial Landmark Detection. 10 2015.

[15] Zhujin Liang, Shengyong Ding, and Liang Lin. Unconstrained Facial Landmark Localization with Backbone-Branches Fully-Convolutional Networks. *arXiv:1507.03409 [cs]*, 1, 7 2015.

[16] Stephen Milborrow and Fred Nicolls. Active Shape Models with SIFT Descriptors and MARS. *Proceedings of the 9th International Conference on Computer Vision Theory and Applications*, (i):380–387, 2014.

[17] Alejandro Newell, Kaiyu Yang, and Jia Deng. Stacked Hourglass Networks for Human Pose Estimation. In *Computer Vision âĂŞ ECCV 2016*, pages 483–499. Springer, Cham, 10 2016.

[18] Deva Ramanan, Xiangxin Zhu, and Deva Ramanan. Face detection, pose estimation, and landmark localization in the wild. *2012 IEEE Conference on Computer Vision and Pattern Recognition*, pages 2879–2886, 6 2012.

[19] Shaoqing Ren, Xudong Cao, Yichen Wei, and Jian Sun. Face Alignment at 3000 FPS via Regressing Local Binary Features. In *2014 IEEE Conference on Computer Vision and Pattern Recognition*, pages 1685–1692. IEEE, 6 2014.

[20] Christos Sagonas, Georgios Tzimiropoulos, Stefanos Zafeiriou, and Maja Pantic. 300 Faces in-the-Wild Challenge: The First Facial Landmark Localization Challenge. In *2013 IEEE International Conference on Computer Vision Workshops*, pages 397–403. IEEE, 12 2013.

[21] G. Storey, R. Jiang, and A. Bouridane. Role for 2D image generated 3D face models in the rehabilitation of facial palsy. *Healthcare Technology Letters*, 4(4), 2017.

[22] Gary Storey, Richard Jiang, and Ahmed Bouridane. Role for 2D image generated 3D face models in the rehabilitation of facial palsy. *Healthcare Technology Letters*, 4(4):145–148, 8 2017.

[23] Yi Sun, Xiaogang Wang, and Xiaoou Tang. Deep Convolutional Network Cascade for Facial Point Detection. In *2013 IEEE Conference on Computer Vision and Pattern Recognition*, pages 3476–3483. IEEE, 6 2013.